# Deep ensemble learning for Alzheimer's disease classification

Ning An
School of Computer and Information
Hefei University of Technology
Hefei, China
ning.g.an@acm.org

Huitong Ding
School of Computer and Information
Hefei University of Technology
Hefei, China
School of Medicine, Boston University
Boston, USA
dinghfut@bu.edu

Jiaoyun Yang[†]
School of Computer and Information
Hefei University of Technology
Hefei, China
jiaoyun@hfut.edu.cn

Rhoda Au
Schools of Medicine & Public Health
Boston University
Boston, USA
rhodaau@bu.edu

Ting Fang Alvin Ang
School of Medicine
Boston University
Boston, USA
alvinang@bu.edu

## ABSTRACT

Ensemble learning use multiple algorithms to obtain better predictive performance than any single one of its constituent algorithms could. With growing popularity of deep learning, researchers have started to ensemble them for various purposes. Few if any, however, has used the deep learning approach as a means to ensemble algorithms. This paper presents a deep ensemble learning framework which aims to harness deep learning algorithms to integrate multisource data and tap the 'wisdom of experts'. At the voting layer, a sparse autoencoder is trained for feature learning to reduce the correlation of attributes and diversify the base classifiers ultimately. At the stacking layer, a nonlinear feature-weighted method based on deep belief networks is proposed to rank the base classifiers which may violate the conditional independence. Neural network is used as meta classifier. At the optimizing layer, under-sampling and threshold-moving are used to cope with cost-sensitive problem. Optimized predictions are obtained based on ensemble of probabilistic predictions by similarity calculation. The proposed deep ensemble learning framework is used for Alzheimer's disease classification. Experiments with the clinical dataset from national Alzheimer's coordinating center demonstrate that the classification accuracy of our proposed framework is 4% better than 6 well-known ensemble approaches as well as the standard stacking algorithm. Adequate coverage of more accurate diagnostic services can be provided by utilizing the wisdom of averaged physicians. This paper points out a new way to boost the primary care of Alzheimer's disease from the view of machine learning.

## KEYWORDS

†Corresponding author
Tel.: +86 13866719116
Fax: +86 055162901990
Email: jiaoyun@hfut.edu.cn
Address: Hefei Tunxi Road 193, Hefei University of Technology, Hefei, 230009, China

Deep learning, Ensemble learning, Stacking, Classification, Alzheimer's disease, Primary care

## 1 INTRODUCTION

Ensemble learning utilizes a group of decision making system which applies various strategies to combine classifiers to improve prediction on new data. Stacking is a well-known approach among the ensembles in which the predictions of a collection of models are given as inputs to a second-level learning algorithm. It has been employed successfully on a wide range of problems, such as chemometrics [1], spam filtering [2], signal processing [3, 4], and healthcare [5]. But the correlation of base classifiers is hard to eliminate. Currently, most methods are focusing on the diversity among the members of a team of classifiers. For example, different learning algorithms and training data sets have been used for this purpose [6, 7]. But few efforts have been made to reduce the correlation of base classifiers in the second-level algorithm of stacking.

The restricted boltzmann machine (RBM) is a representative example of deep learning which has become a major tool in several applications over the last decades, including image recognition [8], bioinformatics [9] and natural language processing [10]. It is a probabilistic model that uses a layer of hidden binary variables or units to model the distribution of a visible layer of variables. As a generative model, it has been used for analyzing different types of data including labeled or unlabeled images [11], and acoustic data [12]. RBM does not require the independent of input components [11]. It is indeed an advantage to fuse the predictions of base classifiers even they might dependent with each other.

Alzheimer's disease (AD) is a chronic neurodegenerative disorder, which makes up more than 60% of all dementia cases [13, 14]. Age is the major risk factor for AD. With a rapidly aging world population, diagnosis services in many middle-income countries strive to meet actual demand and are largely confined to tertiary care hospitals in major population centers [15]. Deep learning with some variants has been used for AD prediction in previous works [16, 17], but lacked the generalization capability needed for application by medical practitioners owing to insufficient data and the inherent physicians' bias clinical judgement. Making full use of limited resources to improve AD diagnostic accuracy poses a severe challenge in improving healthcare. Hence there is an increasing need for new methods that can enhance the primary care of AD.

The diagnosis of AD is generally based on history-taking, clinical presentation and behavioral observations. Specialists working in memory clinics sometimes show surprisingly low levels of diagnostic agreement with each other [18], making it hard to obtain objective and reproducible diagnose. Alternatively, more opinions should be sought from the primary care services because of the lack of AD specialists in many parts of the world. Therefore, it is important to find ways to better leverage the wisdom of experts [19]. Our framework is an effective strategy to assist existing or new health professionals, who have insufficient AD related training, in making clinical diagnosis.

We regard the clinical decision making of physician as a learning algorithm that searches a hypothesis space about AD outcome for the best one. Without sufficient data or expertise, the learning algorithm or physician may derive different AD outcome hypotheses in hypothesis space that all result in the same level of predictive accuracy. By constructing an ensemble of these classifiers or physicians, the algorithm can average decisions and reduce the risk of reliance on the wrong classifier or physician. Many learning algorithms perform local searches for outcome hypothesis that are constrained in local optima. Similarly, physicians may have more expertise in a specific disease and thus their diagnoses are often biased to what they are most familiar with. An ensemble may provide a better approximation to the true unknown outcome than any individual classifier. Wu et al. combined three different classifiers using weighted and unweighted schemes to improve AD prediction [20], but they only use the 11C-PIB PET image data and did not consider the diversity of base classifiers. In other word, the base classifiers may dependent with each other. There have been recent works on how to combine ensemble learning with deep learning systems to achieve greater prediction accuracy [21, 22].

Most of the existing frameworks for AD prediction tend to achieve lower error rates by assuming the same loss for any misclassification. Beheshti developed a novel computer-aided diagnosis system that uses feature ranking and genetic algorithms to analyze structural magnetic resonance imaging data [23]. Using this system, the conversion of mild cognitive impairment (MCI) to AD is predicted. However, different mistakes may lead to significantly different clinical consequences. For example, failing to detect AD has more potentially significant consequence than a false positive prediction. Cost-sensitive learning provides a solution to this problem by considering misclassification costs in the learning process [24].

Although using automated computer tools to facilitate medical analysis and prediction is a promising and important area [25], most existing classification methods only use one individual modality of biomarkers for AD prediction and the data collection process is subject to variability, which may affect the overall classification performance. For example, voxel wise tissue probability, cortical thickness, and hippocampal volumes are all neuroimaging features often used for AD classification [26-28]. There are, however, also a number of biological and/or genetic biomarkers that have been identified as well as being significantly related to increased risk of AD. Actually, different measures provide complementary information, which in combination may significant increase AD prediction performance. The uniform data set (UDS) collected by national Alzheimer's coordinating center (NACC) includes detailed clinical information of participants, such as cognition outcome, neuropsychological test results and family history, as well as neuroimaging indices of neurodegeneration [29]. It is a valuable resource which has promoted a wealth of Alzheimer's disease research findings [30-32].

Based on this multi-dimensional data, we propose a deep ensemble learning framework (DELearning) to leverage clinical expertise of averaged physicians to obtain more accurate AD prediction. It could be used in primary care settings in which there are limited accesses to specialists. DELearning is a three-layer framework with five stages. Firstly, to fuse multi-source data and reduce the correlation of original features, sparse auto-encoder (SAE) is used for feature learning to construct three feature spaces. Secondly, extensive classifiers are built by using different learning algorithms and feature spaces. Multiple hypotheses that can be likened to different physician opinions are generated through this kind of manipulation of training data. Thirdly, a new dataset composed of prediction values of classifiers is fed to a deep belief network (DBN) which uses the stacking method to tackle violations of conditional independence of the base classifiers. Fourthly, three neural networks (NNs) are constructed based on a back-propagation algorithm and several cost sensitive methods, such as under sampling and threshold moving. Finally, probabilistic predictions of these models are mapped in a three-dimensional space. Prototypes of different categories were extracted based on mean values. Discrimination was carried out based on the similarity between individuals and the prototypes.

The contributions of this paper are as follows.

1. We propose a new nonlinear stacking method based on deep learning to cope with the dependence of base classifiers in the second-level learning algorithm.
2. A deep ensemble learning framework is proposed to classify AD outcome. In this framework, base classifiers served as surrogates to physicians with different clinical expertise. DELearning can evaluate different experts and integrate their diagnosis outcome through a contrastive divergence learning procedure [33]. It can utilize both dichotomizing and probabilistic opinions of physicians to make a more accurate diagnosis.

The remainder of this paper is organized as follows. Section 2 presents the learning methods of DELearning. Section 3 discusses the empirical results and some observations. Section 4 presents the conclusion and future work.

## 2 METHODS

In this paper, we focus on two outcomes: probable and possible AD (AD) and non-demented control (NDC). Probable and possible AD is terminology used in all clinical settings [34]. Suppose we are given a group of participants $\{x_1, x_2, ..., x_n\}$, where the sample size is denoted by $n$. They are $n$ independently identically distribution samples from a distribution p($x$, $y$) defined on R, with the label $y \epsilon$ {AD, NDC}. We refer to $X_i \in \{+1, -1\}^d$ as the predictions of $d$ physicians or base classifiers for the subject $x_i$. Each physician or classifier $C_i$ gets $n$ subjects to predict. Subjects will be predicted by all physicians. In this paper, we consider the case where there are physicians with less expertise or weak classifiers and try to address the question of how to obtain high prediction accuracy based on these poorer sources of information. Refer to **Figure 1** for the framework of DELearning which composes of three ensemble layers.

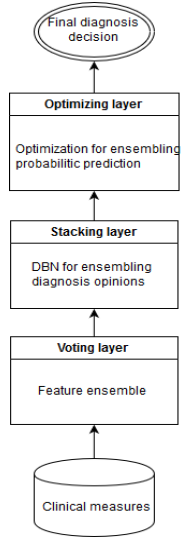

**Figure 1. Framework of DELearning**

**Voting layer**
First, SAE is used for feature learning to fuse multisource sparse data and to construct three feature spaces with reduced correlation of attributes [35]. Next, base classifiers are built by different learning algorithms and feature subsets to generate multiple diverse diagnosis hypotheses. Finally, the participants are quantified by the predictions of these classifiers.

**Stacking layer**
DBN is trained on the quantized participants as the ensemble method to tackle dependence of the base classifiers. As a meta-classifier, DBN combines the predictions of base classifiers in a weighted manner. It is a probabilistic generative model comprised of multiple layers of stochastic and latent variables and restricted boltzmann machine (RBM). Uri Shaham et al. [36] has validated that all classifiers are conditionally independent, is equivalent to a RBM with a single hidden node.

**Optimizing layer**
Three back-propagation NNs are built by cost-sensitive method and assembled to optimize the prediction. $NN_1$ is trained on the prediction of base classifiers with threshold moving. $NN_2$ are initialized by the parameters of trained DBN model and trained on the same dataset as $NN_1$. $NN_3$ is trained on the dataset which distribution is adjusted by over-sampling. Then, we map the probabilistic diagnosis of these NNs in a 3-dimensional space and choose the mean values vector as the prototypes of AD and NDC. The similarities between participant and the 2 prototypes are calculated by Euclidean distance. Finally, the outcome of prototype which is closest to the participant is selected as the final diagnosis.

## 2.1 Datasets

The data comes from the national Alzheimer's coordinating center (NACC) which founded in 1999 and maintains a cumulative database consisting of various types of clinical data such as clinical evaluations, brain MRI imaging and neuropathology. Many researchers have been making use of this resource to get valuable findings [37, 38]. 23,165 samples and 100 attributes are extracted from NACC UDS [29]. There are 6 groups of category variables are selected including medical history (MH), history of hachinski ischemic score (HIS) and cerebrovascular disease (CVD), unified Parkinson's disease rating scale (UPDRS), neuropsychiatric inventory questionnaire (NPIQ), geriatric depression scale (GDS) and functional assessment (FS). Refer to **Table 1** for the details.

**Table 1. Six groups of measures selected from NACC UDS.**

| Groups | Measures |
| --- | --- |
| MH | CVHATT, CVAFIB, CVANGIO, CVBYPASS, CVPACE, CVCHF, CVOTHR, CBSTROKE, CBTIA, CBOTHR, PD, SEIZURES, TRAUMBRF, HYPERTEN, HYPERCHO, DIABETES, B12DEF, THYROID, INCONTU, INCONTF |
| HIS and CVD | ABRUPT, STEPWISE, SOMATIC, EMOT, HXHYPER, HXSTROKE, FOCLSYM, FOCLSIGN, HACHIN, CVDVOG, STROKCOG, CVDIMAG, CVDIMAG1, CVDIMAG2, CVDIMAG3, CVDIMAG4 |
| UPDRS | SPEECH, FACEXP, TRESTFAC, TRESTRHD, TRESTLHD, TRESTRFT, TRESTLFT, TRACTRHD, TRACTLHD, RIGDNECK, RIGDUPRT, RIGDUPLF, RIGDLORT, RIGDLOLF, TAPSRT, TAPSLF, HANDMOVR, HANDMOVL, HANDALTR, HANDALTL, LEGRT, LEGLF, ARISING, POSTURE, GAIT, POSSTAB, BRADYKIN |
| NPIQ | DEL, HALL, AGIT, DEPD, ANX, ELAT, APA, DISN, IRR, MOT, NITE, APP. |
| GDS | SATIS, DROPACT, EMPTY, BORED, SPIRITS, AFRAID, HAPPY, HELPLESS, STAYHOME, MEMPROB, WONDRFUL, WRTHLESS, ENERGY, HOPELESS, BETTER |
| FS | BILLS, TAXES, SHOPPING, GAMES, STOVE, MEALPREP, EVENTS, PAYATTN, REMDATES, TRAVEL |
| Total number | sample: 23165; outcome: 2 ; measure: 100 |

## 2.2 Voting layer

Due to heterogeneous nature of 6 groups of attributes, single classifier has difficulty in sufficiently leveraging multisource information to obtain a satisfied performance on AD classification no matter the amount of available data. More specifically, the clinical decision boundary that discriminate participants from different outcomes may linear for some attributes while non-linear for another part. It may lie outside the space of functions that can be implemented by the chosen classifier. Even though the single classifier could achieve satisfied classification performance on the available data, it might not generalize for another data sources.

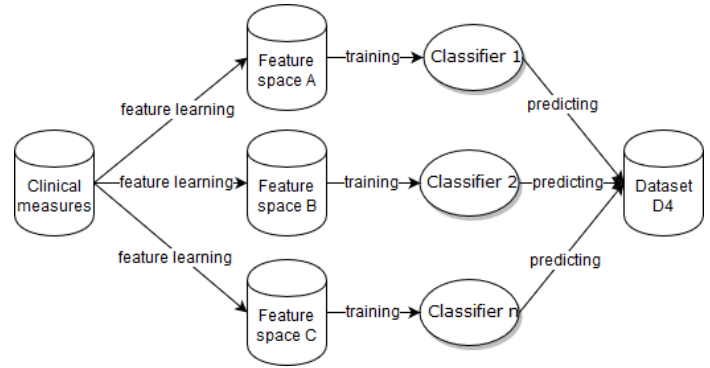

**Figure 2. Voting layer of DELearning.**

As shown in **Figure 2**, DELearning constructs different feature spaces and trains multiple classifiers with diverse learning algorithms on it to increase the generalization ability and diversify the decision boundaries. These classifiers consist bayes network, filtered classifier, hoeffding tree, iterative classifier optimizer, J48, logistic regression, logitboost, multiclass classifier, random committee, random forest, random subSpace, REPTree, AdaBoostM1, multilayer perception, naïve bayes, stacking, voted perceptron, vote. Here we illustrate three of them.

Logistic regression is a well-behaved classification algorithm especially when the features to be studied can be treated as roughly linear or the problem is linearly separable [39]. Feature engineering technologies can transform most non-linear features into linear ones. This method is also robust to noise and can avoid overfitting. It even can be applied in the field of feature selection by using $L_1$ or $L_2$ regularization.

Tree ensembles are combination of a bunch of decision trees [40]. One dominant advantage of tree ensemble is that they do not presume linear features for data. So they are quite suitable for handling categorical features.

The Bayes classifier is probabilistic classifier which applying Bayes' theorem with strong independence assumptions between the features [41]. If the conditional independence assumption is true, the Bayes classifier will converge faster than discriminative models e.g logistic regression. So it needs less training data. And even if the assumption fails, the Bayes classifier still performs beyond expectations in testing.

Various indicators have been utilized for quantitative assessment of diversity. Q-statistic is a measure to assess the similarity of two classifiers' predictions [42]. It is formulated as

$$Q_{i,j} = \frac{N_{11}N_{00} - N_{01}N_{10}}{N_{11}N_{00} + N_{01}N_{10}} \quad (1)$$

where $N_{ab}$ is the number of participants which has outcome $a$ but recognized as outcome $b$.

We define a discrete random variable $E$ taking values from $\{0, \frac{1}{L}, \frac{2}{L}, ..., \frac{L-1}{L}, 1\}$. It denotes the proportion of classifiers in set $D$ that correctly classify participant $x$ randomly selected from the cohort. To estimate the probability mass function of $E$, we run the $L$ classifiers in $D$ on the data set. We can capture the distribution shape by using the variance of $E$. The difficulty $\theta$ is defined as variances of $E$ [42].

We treat the trained classifiers as physicians from different fields with different clinical expertise. For a participant, the prediction of the classifier is the diagnosis of corresponding physician.

In addition to the various learning algorithms of base classifiers, resampling of the training data is another way to increase the diversity of the ensemble. The clinical measures extracted from NACC are sparse, which affects the performance of most classifiers. With the purpose of improving the diversity of stacking layer and reduce the sparsity of data, SAE is used in voting layers to automatically learn different features space defined by the activations of its hidden nodes. Refer to **Figure 3** for the details of SAE structure. The most important advantage of feature learning by SAE is that the correlation of transformed features is greatly reduced. Thereby, trained on these feature spaces, the diversity of base classifiers is increased to a greater extent. Suppose we have only an unlabeled sample set $x=\{x_1, x_2, ..., x_n\}$, where $x_i$ belongs to $R^n$. An auto-encoder neural network is an unsupervised learning algorithm that applies backpropagation, setting the target values to be equal to the inputs.

The SAE tries to learn an approximation to the identity function, so that output $\bar{x}$ is similar to $x$. By constraining the network, such as limiting the number of hidden units, we can discover interesting structures about the data. Informally, assuming a sigmoid function, we will regard a neuron as being active if its output value is close to 1, or as being inactive if its output value is close to 0.

Let $a_j^{(2)}$ is the activation of hidden unit $j$ in the SAE. This notation does not clearly show the input value of $x$ that led to that activation. Thus, when the network is given a specific input $x$, we will modify $a_j^{(2)}$ to better denote the activation of this hidden unit as $a_j^{(2)}(x)$. Further, let

$$\hat{\rho}_j = \frac{1}{m}\sum_{i=1}^{m}[a_j^{(2)}(x^{(i)})] \quad (2)$$

be the average activation of hidden unit $j$ over the training set. We would like to enforce the constraint

$$\hat{\rho}_j = \rho \quad (3)$$

where $\rho$ is sparsity parameter. We compare various commonly used values and chose $\rho = 0.05$ in DELearning, as shown in **Figure 7**.

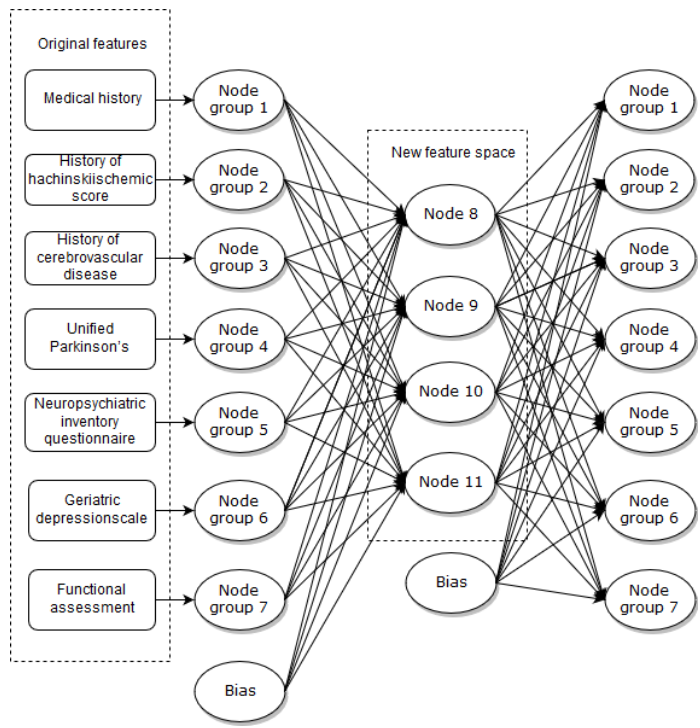

**Figure 3. The SAE for feature learning. Features are extracted by sequentially propagating 7 groups of clinical measures for each participant through the hidden layer of the SAE.**

## 2.3 Stacking layer

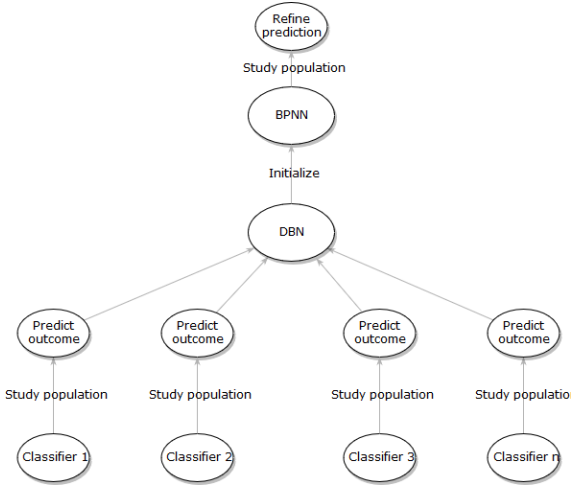

**Figure 4. Stacking layer of DELearning**

Stacking is an ensemble technique in which the predictions of a group of classifiers are given as inputs to a second-level learning algorithm. Then, the second-level algorithm is trained to obtain an optimization decision. RBM is a probabilistic model that uses unobserved random variables to model the distribution of observed data. Typically, an RBM is trained in an unsupervised manner to model the distribution of the inputs. The most outstanding strength of RBM is that the hidden units are conditionally independent given the visible units [11]. We can reduce the correlation of prediction of base classifiers or physicians by the data transformation in RBM. As shown in **Figure 4**, we propose a stacking method based on deep learning which can be achieved by three steps. First, we train the DBN with the prediction of base classifiers. Then, the accumulation weight of each input nodes is used to rank the base classifiers or physicians and initialize the same configuration neural network. Finally, this network is trained to obtain the preliminary probabilistic prediction.

## 2.4 Optimizing layer

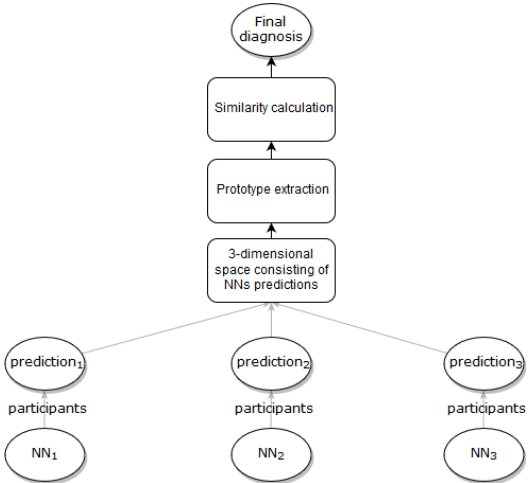

**Figure 5. Optimizing layer of DELearning**

Two challenges in AD primary care are that the amount of patients is fewer than healthy people and the cost of missed diagnosis is greater than that of misdiagnosis. Cost-sensitive learning is a suitable tool for learning from imbalanced dataset and learning when costs are unequal or unknown. DELearning adapts over-sampling, threshold moving and soft ensemble to train NNs and ensemble their probabilistic predictions.

Assuming there are $W$ outcomes for a disease, $N_i$ participants in $i$-th outcome. Let $Cost[i, j]$ ($i, j \in \{1, 2, ..., W\}$) be the cost of diagnose a participant that is $i$-th outcome as the $w$-th outcome ($Cost[i, i] = 0$). $Cost[i]$ is the cost of the $i$-th outcome. We presume that the outcome is ordered. In other word, if $i < j$ then $Cost[i] < Cost[j]$, or $Cost[i] = Cost[j]$ and $N_i \geq N_j$. $Cost[i]$ is generally obtained from $Cost[i, j]$ as follows, $Cost[i] = \sum_{j=1}^{W} Cost[i, j]$ [24]. In DELearning, $W=2$ represents there are two outcomes, AD and NDC.

The $k$-th outcome will have $N_k^*$ individuals after resampling according to [24].

$$N_k^* = \left\lfloor \frac{Cost[k]}{Cost[\lambda]} N_\lambda \right\rfloor \quad (4)$$

The $\lambda$-th outcome which has the fewest participants to be resampling is identified.

$$\lambda = \arg\min_j \frac{\frac{Cost[j]}{\min_c Cost[c]} N_{\arg\min_c Cost[c]}}{N_j} \quad (5)$$

If $N_k^* > N_k$, we sample $N_k^* - N_k$ participants in $k$-th outcome randomly with replacement.

Besides over-sampling, we adjust the threshold of output unit in NN toward inexpensive outcome in order that participant with higher misclassification costs are difficult to be neglected.

Specifically, let $O_i$ be the real-valued outputs of neural network, where $i \in \{1, 2, ..., W\}$. In DELearning the prediction outcome is $O_i^*$ which is calculated according to equation 6 [24].

$$O_i^* = \eta \sum_{c=1}^{C} O_i Cost[i, c] \quad (6)$$

There are three types of cost matrices defined in [24]. In our AD classification task, these cost matrices are the same and fall into one of the three types. Without loss of generality, we impose the following unity condition and design a cost matrix as follows:

$$1.0 \leq Cost[i, j] = H_i \leq 10.0 \quad (7)$$

for each $j \neq i$, $Cost[i] = H_i$. At least one $H_i = 1.0$. Under this condition, various cost matrices are compared in terms of Geometric mean of NNs.

Since under-sampling discards potentially useful training samples, the performance of the resulting classifier may be degraded. DELearning uses over-sampling for training NN in optimizing layer. The effectiveness of over-sampling has been proved in former study [43].

After training cost-sensitive NNs through over-sampling and threshold-moving, we combine NNs into an ensemble. The process can be found in **Figure 5**. We map the participants into the three-dimensional space composed by probabilistic predictions of NNs. The mean values are chose as the prototypes of AD and NDC. As predictions of new participant the three NNs come in, we mapped them in the three-dimensional space and computed the Euclidean distance between them and the prototypes of AD and NDC. The category with the smallest distance was selected as the final optimized prediction result.

The DELearning algorithm is shown in **Table 2**.

**Table 2. Proposed DELearning Algorithm.**

| |
|---|
| Input: Study population D with clinical measures and outcomes; |
| Output: Prediction of AD outcome |
| **Voting Layer** |
| 1 Normalize the original data set D, get $D_1$ |
| 2 Train SAEs on $D_1$ |
| 3 Derive the optimal structure of SAE based on lowest mean square error (MSE) |
| 4 Construct feature spaces, $D_2$ and $D_3$, by activation values of hidden layer in SAE |
| 5 Train various classifiers on $D_1$, $D_2$ and $D_3$, such as Bayes network, naïve Bayes, multilayer perceptron |
| 6 Generate data $D_4$, which is composed of the predictions for all participants outcome of all classifiers on $D_1$, $D_2$ and $D_3$ |
| **Stacking Layer** |
| 7 Train DBN on $D_4$ with minimum reconstruction error |
| 8 Evaluate classifiers via accumulating weights of input nodes to hidden nodes in DBN. |
| 9 Use back-propagation neural network as meta-classifier to search a blended prediction function. |
| **Optimizing Layer** |
| 10 Initialize neural networks $NN_1$, $NN_2$, $NN_3$ with the parameters of DBN |
| 11 Train $NN_1$ on $D_4$ |
| 12 Identify the cost matrixes in $NN_1$ with highest geometric mean |
| 13 Accumulate output of $NN_1$ with the cost and predict strategy is set to biggest value |
| 14 Train $NN_2$ using the following steps: Initialize $D_5$ by $D_4$ Resample ($N_{NDC}-N_{AD}$) participants from $D_5$ which is AD and put them in $D_5$ Train $NN_2$ using $D_5$ |
| 15 Train $NN_3$ using $D_4$ |
| 16 Produce probabilistic outputs $P_1$ of $NN_1$ and multiply them with misclassification costs |
| 17 Produce probabilistic outputs $P_2$ of $NN_2$ |
| 18 Produce probabilistic outputs $P_3$ of $NN_3$ |
| 19 Formulate study population with ($P_1$, $P_2$, $P_3$) |
| 20 Generate prototypes for 2 outcomes |
| 21 Calculate the similarity between participants and prototypes |
| 22 Treat the outcome of prototype which is most similar with participant as the final diagnosis; if the similarity is equal for 2 prototypes, then the final diagnosis is the outcome with the biggest misclassification cost |

## 2.4 Performance measures

A confusion matrix that contains the actual outcome and predicted outcome is used to evaluate the performance of AD classification. **Table 3** presents an example of confusion matrix for AD classification with two outcomes. TP is the number of AD patients that are correctly classified as AD. FP is the number of NDC participants that are diagnosis as AD. FN is the number of AD patients that are incorrectly classified as NDC. TN is the number of NDC participants that are classified correctly. We use the following four measures to evaluate the AD classifiers. The higher accuracy, precision, recall rate and F-measure, the better the constructed classifier.

**Table 3: Confusion Matrix for AD Prediction**

| | | Diagnosis outcome | |
|---|---|---|---|
| | | AD | NDC |
| True outcome | AD | TP | FN |
| | NDC | FP | TN |

Accuracy is the probability of correctly diagnosis of the outcome for each participant. It is formulated as the follows.

$$\text{Accuracy} = \frac{TP + TN}{TP + FP + FN + TN} \qquad (8)$$

Precision denotes the proportion of predicted AD cases that are correctly real AD.

$$\text{Precision} = \frac{TP}{TP + FP} \qquad (9)$$

Recall is the proportion of AD participants that are correctly classified. Its desirable feature is that it reflects how many of the relevant participants predicted positive rule of the classifier picks up. In a medical context, recall is regarded as primary measure, as the aim is to identify all real positive cases.

$$\text{Recall} = \frac{TP}{TP + FN} \qquad (10)$$

F-measure provides a way to integrate precision and recall into a single measure. It takes value from 0 to 1. And if the F-measure of a classifier equals to 1, the classifiers can correctly classify all participants.

$$\text{F-measure} = 2 \times \frac{\text{Precision} \times \text{Recall}}{\text{Precision} + \text{Recall}} \qquad (11)$$

## 3 RESULTS AND DISCUSSION

## 3.1 Feature learning

DELearning utilizes SAE as a feature learning method to obtain more discriminative features compared with the original set. In order to determine the optimal dimension of transformed space and reduce data sparsity, we trained three-layer SAE models with 100 input units and 10, 20, 30 hidden units respectively. The activation function was sigmoid. Visible biases and weights were initialized to zero and random numbers sampled from a zero-mean normal distribution with standard deviation 0. Momentum was set to 0.5. The model was trained for 200 epochs. As shown in **Figure 6**, the mean square error (MSE) of models with 20, 30 hidden units are much smaller than the model with 10 hidden units. Thus we determine the number of hidden units of SAEs is 20, 30 for the transformation of feature space.

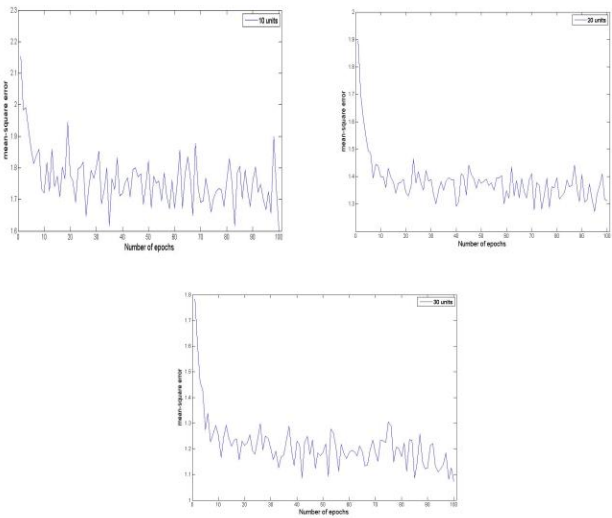

**Figure 6. The comparison of SAE with 10, 20, 30 hidden units. The x-axis is the number of epoch. The y-axis is the MSE in corresponding epoch.**

As shown in **Figure 7,** we compared the MSE of SAEs with to determine the sparsity parameter ρ with commonly used values 0.01, 0.05, 0.1, 0.15. The best performance was observed at ρ = 0.05. So we set ρ =0.05 in DELearning.

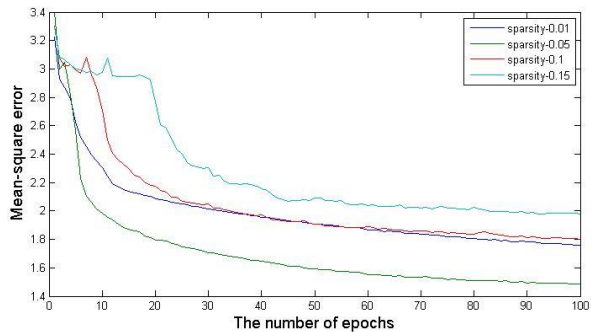

**Figure 7. The MSEs of SAEs with sparsity represented by colored line.**

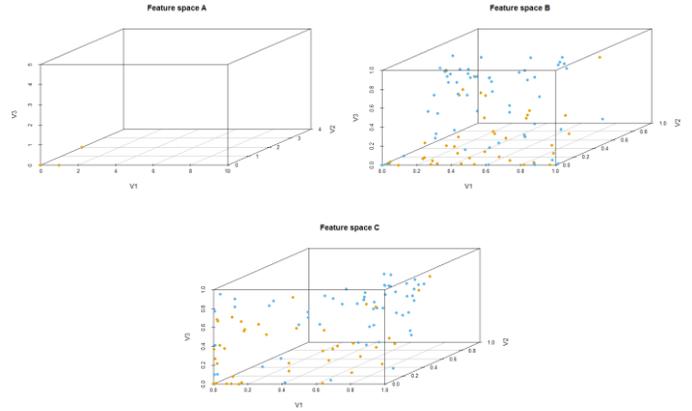

**Figure 8. Visualization of 100 participants in 3 dimensions selected in original space (a), feature space learned by SAE with 20 hidden units (b) and feature space learned by SAE with 30 hidden units (c). The circle represents participant. The color of circle indicates the outcomes, orange for AD, blue for NDC.**

The samples after feature transformation are easier to distinguish. **Figure 8** is the visualization of 100 randomly-selected participants in 3 randomly-selected attributes of 3 feature spaces. In the original space, most participants with different outcomes are overlap with each other. With the transformed features, the boundary between AD and NDC is relatively obvious, which intuitively indicates that the transformed feature spaces are not only reduced but also more adept at representing the two outcome groups.

**Figure 9** shows the correlation matrices of features in different spaces we constructed. It demonstrates that our framework reduces the dependence of attributes in the original dataset. Consequently, new feature spaces tend to approximate low to no collinearity. It helps to improve the performance of base classifiers such as accuracy and speed.

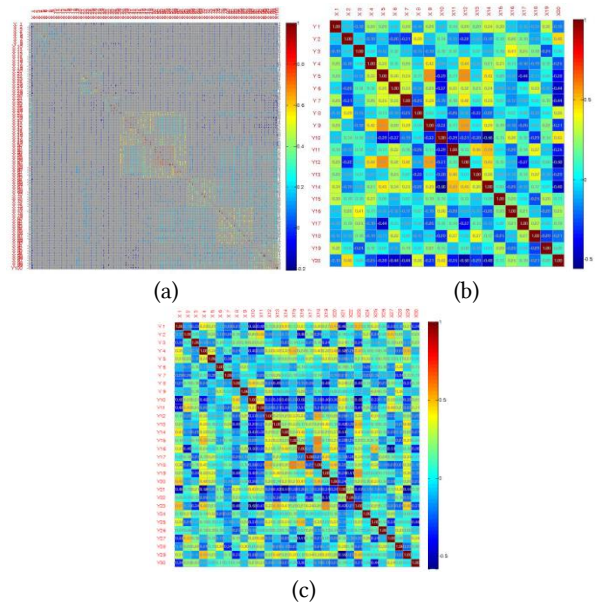

**Figure 9.** Correlation matrix of three feature spaces, original space (a), 20-dimensional feature space transformed by SAE (b) and 30-dimensional feature space transformed by SAE (c). The shaded rectangular tiling represents the correlation value of corresponding two attributes. It uses a color scale ranging from bule (low correlation value) to red (high correlation value). It clearly indicates that most of the features in original space are not conditionally independent of each other. The features in 20-dimensional and 30-dimensional feature spaces are approximately uncorrelated.

### 3.2 Construction and evaluation of classifiers

In this section, we show the AD classification performance of base classifiers on original UDC and two transformed datasets. Here are some details of the classifiers. The structure learning method of our Bayes network is hill climbing search for optimal Bayes score. Maximum likelihood estimates method for the parameter learning with obtained structure. Filtered classifier is a combination of J48 decision tree and MDL discretization method [44]. The minimum number of participants in the leaf is set to 2. 3-fold of participants is used for pruning with the confidence factor is 0.25. The rest is used for growing the tree. Hoeffding tree is an incremental, anytime decision tree induction algorithm. The splitting criterion is gini index. The leaf prediction strategy is set to majority class. REPTree is a decision tree using information gain and prune with reduced-error pruning with backfitting. Random subspace is consists of multiple REPTree constructed systematically by randomly selecting 50% components of the all features. The activation function of Multilayer Perception is sigmod function. The number of hidden units is set to 51, 11 and 16 for the models in 3 different feature spaces. The learning rate is set to 0.3. Momentum is 0.2.

**Table 4** to **Table 7** show that the performance of base classifiers in the reduced feature spaces, namely 20 and 30 dimensional spaces learned by SAE, are comparable with the original feature space. It indicates that our feature learning method not only greatly reduces the correlation of clinical measures but also retains the useful information. At the same time, physicians in different fields have been imitated by these classifiers.

**Table 4.** Accuracy of base classifiers in three feature spaces, normalized original space (A), 20-dimensional space learned by SAE (B) and 30-dimensional space learned by SAE (C).

| Classifier | Feature space | | |
|---|---|---|---|
| | A | B | C |
| Bayes Nets | 75.6% | 71.4% | 72.6% |
| Filtered classifier | 79.2% | 75.4% | 75.3% |
| Hoeffding tree | 78.5% | 76.0% | 74.7% |
| Iterativeclassifieroptimizer | 79.9% | 75.3% | 76.3% |
| J48 | 79.1% | 75.1% | 75.8% |
| Logistic Regression | 80.4% | 77.2% | 78.0% |
| Logit Boost | 79.9% | 75.3% | 76.3% |
| Random committee | 82.7% | 75.6% | 75.4% |
| Random forest | 81.7% | 83.9% | 78.4% |
| Random SubSpace | 81.8% | 76.1% | 77.4% |
| REPTree | 80.4% | 76.0% | 75.9% |
| AdaBoostM1 | 76.2% | 74.4% | 75.5% |
| Multilayer Perception | 80.5% | 79.7% | 80.1% |
| Naïve Bayes | 72.1% | 71.5% | 73.6% |
| Stacking | 63.1% | 62.6% | 63.1% |
| Voted Perceptron | 78.5% | 77.4% | 76.4% |

**Table 5.** Precision of base classifiers in three feature spaces, normalized original space (A), 20-dimensional space learned by SAE (B) and 30-dimensional space learned by SAE (C).

| Classifier | Feature space | | |
|---|---|---|---|
| | A | B | C |
| Bayes Nets | 76.1% | 73.9% | 73.1% |
| Filtered classifier | 79.1% | 75.5% | 75.1% |
| Hoeffding tree | 79.1% | 76.1% | 74.6% |
| Iterativeclassifieroptimizer | 79.8% | 76.1% | 76.4% |
| J48 | 78.9% | 74.7% | 76.2% |
| Logistic Regression | 80.2% | 77.3% | 78.0% |
| Logit Boost | 80.1% | 76.1% | 76.4% |
| Random committee | 82.7% | 75.2% | 75.1% |
| Random forest | 80.1% | 80.9% | 78.1% |
| Random SubSpace | 81.8% | 75.8% | 77.2% |
| REPTree | 80.4% | 76.3% | 76.0% |
| AdaBoostM1 | 78.5% | 77.1% | 75.8% |
| Multilayer Perception | 80.3% | 79.5% | 78.3% |
| Naïve Bayes | 71.5% | 73.8% | 75.4% |
| Stacking | 60.2% | 61.3% | 60.5% |
| Voted Perceptron | 78.3% | 77.5% | 78.1% |

**Table 6.** Recall rate of base classifiers in three feature spaces, normalized original space (A), 20-dimensional space learned by SAE (B) and 30-dimensional space learned by SAE (C).

| Classifier | Feature space | | |
|---|---|---|---|
| | A | B | C |
| Bayes Nets | 75.6% | 71.4% | 72.3% |
| Filtered classifier | 79.3% | 75.4% | 75.4% |
| Hoeffding tree | 78.5% | 76.0% | 74.7% |
| Iterativeclassifieroptimizer | 79.9% | 75.4% | 76.3% |
| J48 | 79.1% | 75.1% | 75.8% |
| Logistic Regression | 80.5% | 77.3% | 78.1% |
| Logit Boost | 79.9% | 75.4% | 76.3% |
| Random committee | 82.8% | 75.6% | 75.5% |
| Random forest | 82.8% | 84.9% | 77.5% |
| Random SubSpace | 81.9% | 76.2% | 77.4% |
| REPTree | 80.4% | 76.0% | 75.9% |
| AdaBoostM1 | 76.1% | 74.5% | 75.5% |
| Multilayer Perception | 80.4% | 79.8% | 79.9% |
| Naïve Bayes | 72.2% | 71.6% | 73.6% |
| Stacking | 64.2% | 62.7% | 62.8% |
| Voted Perceptron | 78.0% | 77.4% | 77.9% |

**Table 7.** F-measure of base classifiers in three feature spaces, normalized original space (A), 20-dimensional space learned by SAE (B) and 30-dimensional space learned by SAE (C).

| Classifier | Feature space | | |
|---|---|---|---|
| | A | B | C |
| Bayes Nets | 75.8% | 71.9% | 72.1% |
| Filtered classifier | 79.2% | 75.4% | 75.2% |
| Hoeffding tree | 78.7% | 76.0% | 73.5% |
| Iterativeclassifieroptimizer | 79.9% | 75.6% | 76.4% |
| J48 | 79.0% | 74.7% | 76.0% |
| Logistic Regression | 80.2% | 77.3% | 78.0% |
| Logit Boost | 79.9% | 75.6% | 76.4% |
| Random committee | 82.7% | 75.3% | 75.1% |
| Random forest | 82.7% | 82.9% | 78.4% |
| Random SubSpace | 81.8% | 75.8% | 77.3% |
| REPTree | 80.4% | 76.1% | 75.9% |
| AdaBoostM1 | 76.1% | 74.9% | 75.6% |
| Multilayer Perception | 80.3% | 79.5% | 79.8% |
| Naïve Bayes | 71.4% | 72.0% | 74.0% |
| Stacking | 62.2% | 60.3% | 60.6% |
| Voted Perceptron | 78.1% | 77.5% | 77.8% |

Through the different learning algorithms and feature data, the diversity of these base classifiers or physicians has been maximized. Here is an example to validate it.

**Table 8** is the classification results of logistic regression (LR) and REPTree (RT) in the 20-dimensional feature space. $N_{11}$ and $N_{00}$ are the number of participants which are correctly and wrongly recognized by these two classifiers. $N_{10}$ is the number of participants who were correctly diagnosed by LR but wrongly diagnosed by RT, and $N_{01}$ vice versa. The $Q$ statistics of LR and RT is 0.86<1. This shows that these two classifiers tend to recognize the same individuals correctly.

**Table 8. A 2×2 table of the relationship between logistic regression and REPTree in AD classification on 20-dimensional feature space.**

| | RT correct(1) | RT wrong(0) |
|---|---|---|
| LR correct(1) | $N_{11}$(7535) | $N_{10}$(1102) |
| LR wrong(0) | $N_{01}$(958) | $N_{00}$(1905) |

$$Q_{LR,RT} = \frac{N_{11}N_{00} - N_{01}N_{10}}{N_{11}N_{00} + N_{01}N_{10}} = \frac{14354175 - 1055716}{14354175 + 1055716} = \frac{13298459}{15409891} \approx 0.86$$

Let X is a discrete random variable taking values in $\{0, \frac{1}{L}, \frac{2}{L}, ..., \frac{L-1}{L}, 1\}$ and representing the number of base classifiers that correctly identify a participant. The variance of X is a measure of diversity based on the distribution of difficulty. **Figure 10(a)** shows that all classifiers in the group made the correct predictions for around 4500 samples in the test dataset. All base classifiers which are trained by different learning algorithms and feature spaces achieved relatively high performance. Diverse groups of classifiers will have smaller variance of X. Here, X = 1.1751e+06. As shown in **Figure 10(b)**, after using our classifier ranking method in stacking layer, we observed that the base classifiers No. 3 and 5, have the highest score whose prediction accuracy are concurrently higher. Therefore, DELearning can evaluate classifiers or physicians automatically.

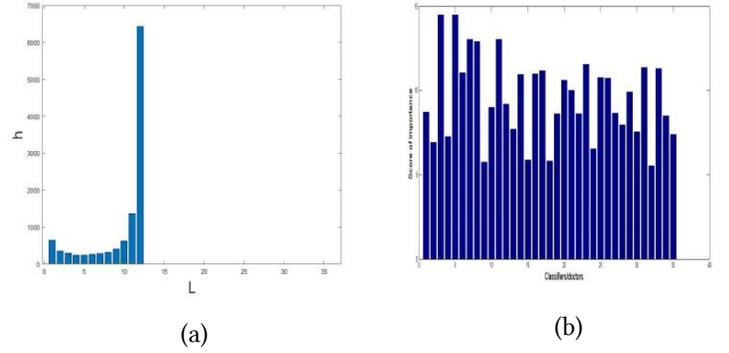

(a) (b)

**Figure 10.** (a) Patterns of difficulty θ for classifiers group with $L$ = 35, $N$ = 11500 and p> 0.7. The histograms show the number of samples which are correctly diagnosis by i of the 35 classifiers. The x-axis is the number of correct classifiers. (b) The important score of base classifiers calculated by DELearning. The x-axis indicates classifier number. The y-axis is the score of corresponding classifier.

### 3.3 Stacking by deep learning

We formulated each participant as a vector composing of the predictions of these base classifiers. DBN was trained to ensemble these opinions from base classifiers with a greedy layer-wise unsupervised method in a mini-batch size of 100 cases [45]. Visible biases and weights were initialized to 0 and random numbers obeyed a zero-mean normal distribution with standard deviation 0.01. Momentum was set as 0.5. We ran 100 epochs with a learning rate of 0.002. At present, there is no golden rule for the selection of optimal number of hidden units. It depends on the types and structures of datasets. With a few units, the training of networks can be speed up but result in poor performance. Numerous units may cause over-fitting and slow learning process. Here, we compared different settings of number of hidden units from 3 to 9. Refer to **Figure 11** for the details. The DBN with 6 hidden units has the lowest error and is selected as the best . The pre-trained DBN were used to initialize artificial neural networks. The other hyperparameters were derived from the process of DBN fine-tuning.

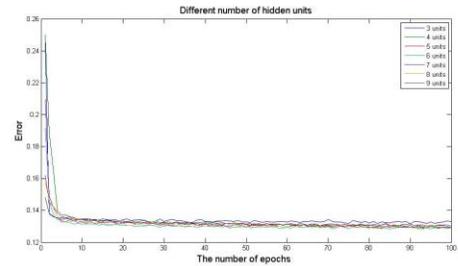

**Figure 11. Errors with different sizes of hidden layer**

### 3.4 Optimization for the final decision

We compared 9 cost matrixes with integer values between 1.0 and 10.0. Each matrix was required to have at least one non-diagonal element to be equal to 1. From **Figure 12** we can see that Geometric mean of cost 2 is highest; the cost matrix we used are shown in **Table 9**.

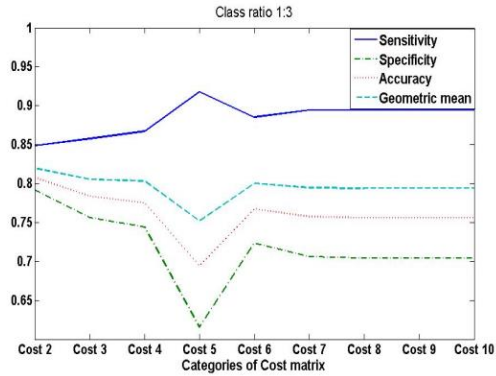

**Figure 12. The performance of NNs with different cost matrixes.**

**Table 9. Cost matrix in DELearning.**

|  | AD | NDC |
|---|---|---|
| AD | 0 | 2 |
| NDC | 1 | 0 |

Specifically, after all the parameters were determined in DELearning, we compared it with 6 ensemble learning methods, including Logiboost [46], Bagging [47], Random forest [48], AdaBoostM1 [49], Stacking [50], Vote [51]. Four algorithms were used to generate individual committees including ZeroR, decision table, Naïve Bayes and multilayer perception. As shown in **Figure 13,** the prediction accuracy of DELearning improved by more than 4% compared with these other approaches on the NACC dataset.

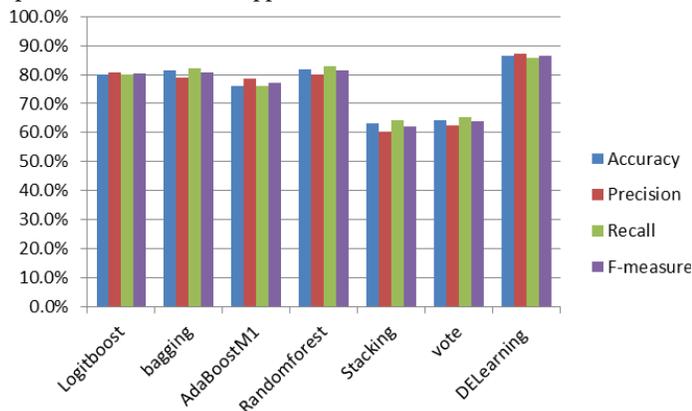

**Figure 13. The performance comparison of DELearning with 6 representative ensemble learning methods on NACC UDC.**

## 4  CONCLUSIONS

The unsurmountable challenge of Alzheimer's disease classification leads us to leverage the wisdom of experts and integrate multisource data to come up with better outcome prediction modality that could be used in primary care. In this paper, we propose DELearning, a three-layer framework for AD classification that uses the deep learning approach to ensemble at each layer.

Using the clinical measures from NACC UDC, we compared the performance of DELearning with 6 representative ensemble learning methods. The experimental results show that DELearning outperforms the other methods in terms of AD prediction accuracy. It provides a data-driven solution to aid AD primary care, particularly where access to AD expertise is limited.

DELearning can also be applied to other scenarios including medical image tagging where it may be not feasible or too expensive to obtain objective and reliable labels. DELearning can collect subjective labels from multiple experts or annotators and find meaningful yet hidden labels.

## ACKNOWLEDGMENTS

This work was supported in part by the National Key R&D Program of China under Grant No. 2018YFB1003204, National Natural Science Foundation of China under Grant No. 71661167004, No. 61502135, the Anhui Key Project of Research and Development Plan under Grant No. 1704e1002221, the Foshan Science and Technology Innovation Project under Grant No. 2015IT100095, and the Program of Introducing Talents of Discipline to Universities ("111 Program") under Grant No. B14025. The NACC database is funded by NIA/NIH Grant U01 AG016976. NACC data are contributed by the NIA-funded ADCs: P30 AG019610 (PI Eric Reiman, MD), P30 AG013846 (PI Neil Kowall, MD), P50 AG008702 (PI Scott Small, MD), P50 AG025688 (PI Allan Levey, MD, PhD), P50 AG047266 (PI Todd Golde, MD, PhD), P30 AG010133 (PI Andrew Saykin, PsyD), P50 AG005146 (PI Marilyn Albert, PhD), P50 AG005134 (PI Bradley Hyman, MD, PhD), P50 AG016574 (PI Ronald Petersen, MD, PhD), P50 AG005138 (PI Mary Sano, PhD), P30 AG008051 (PI Thomas Wisniewski, MD), P30 AG013854 (PI M. Marsel Mesulam, MD), P30 AG008017 (PI Jeffrey Kaye, MD), P30 AG010161 (PI David Bennett, MD), P50 AG047366 (PI Victor Henderson, MD, MS), P30 AG010129 (PI Charles DeCarli, MD), P50 AG016573 (PI Frank LaFerla, PhD), P50 AG005131 (PI James Brewer, MD, PhD), P50 AG023501 (PI Bruce Miller, MD), P30 AG035982 (PI Russell Swerdlow, MD), P30 AG028383 (PI Linda Van Eldik, PhD), P30 AG053760 (PI Henry Paulson, MD, PhD), P30 AG010124 (PI John Trojanowski, MD, PhD), P50 AG005133 (PI Oscar Lopez, MD), P50 AG005142 (PI Helena Chui, MD), P30 AG012300 (PI Roger Rosenberg, MD), P30 AG049638 (PI Suzanne Craft, PhD), P50 AG005136 (PI Thomas Grabowski, MD), P50 AG033514 (PI Sanjay Asthana, MD, FRCP), P50 AG005681 (PI John Morris, MD), P50 AG047270 (PI Stephen Strittmatter, MD, PhD).